\documentclass[letterpaper, 10 pt, conference]{ieeeconf}  

\IEEEoverridecommandlockouts                              

\let\labelindent\relax

\usepackage{graphics} 
\usepackage{epsfig} 
\usepackage{mathptmx} 
\usepackage{times} 
\usepackage{amsmath} 
\usepackage{amssymb}  
\usepackage{algorithm}
\usepackage{algpseudocode}
\usepackage{multirow}
\usepackage{booktabs}
\usepackage{caption}
\usepackage{subcaption}
\usepackage{enumitem}
\usepackage[usenames,dvipsnames]{color}
\usepackage[noadjust]{cite}

\title{\LARGE \bf
AnyPose: Anytime 3D Human Pose Forecasting via \\Neural Ordinary Differential Equations
}

\author{Zixing Wang and Ahmed H. Qureshi
\thanks{The authors are with the Department of Computer Science, Purdue University, IN, USA.
        {\tt\small \{wang5389, ahqureshi\}@purdue.edu}}%
}

\begin{document}

\maketitle
\thispagestyle{empty}
\pagestyle{empty}

\begin{abstract}

Anytime 3D human pose forecasting is crucial to synchronous real-world human-machine interaction, where the term ``anytime" corresponds to predicting human pose at any real-valued time step. However, to the best of our knowledge, all the existing methods in human pose forecasting perform predictions at preset, discrete time intervals. Therefore, we introduce AnyPose, a lightweight continuous-time neural architecture that models human behavior dynamics with neural ordinary differential equations. We validate our framework on the Human3.6M, AMASS, and 3DPW dataset and conduct a series of comprehensive analyses towards comparison with existing methods and the intersection of human pose and neural ordinary differential equations. Our results demonstrate that AnyPose exhibits high-performance accuracy in predicting future poses and takes significantly lower computational time than traditional methods in solving anytime prediction tasks.

\end{abstract}

\section{Introduction}
\label{sec:intro}
Human 3D pose forecasting from past pose sequences is essential to various applications especially involving human-machine interactions (HMI), such as assistive robots~\cite{laplaza2021attention} and autonomous vehicles~\cite{wang2020estimating, du2020unsupervised}. A future estimate of human poses at any given time can improve the decision-making of such systems, resulting in a compliant, natural, and informed HMI. Although several human-pose forecasting methods~\cite{sofianos2021space, mao2020history, mao2019learning, martinez2017human} exist with sequential deep learning-based models leading the performance race, none of these approaches solve the challenge of ``anytime" spatio-temporal pose forecasting. The objective of ``anytime'' pose forecasting is to estimate the human pose at any future time step, requiring continuous-time forecasting of spatio-temporal human poses.

Anytime pose forecasting is crucial to realizing compliant HMI, especially in situations with system latency or scenarios demanding high fidelity motion analysis~\cite{laplaza2021attention, butepage2018anticipating, lang2017object, nemlekar2019object}. Instead of full action trajectories, such tasks are in need of fast and precise pose predictions at one or several designated timesteps. Although previous methods achieved great success in accuracy, the time cost merely meets the requirement for smooth HMI. Furthermore, traditional methods can indirectly achieve anytime forecasting either by interpolating the poses predicted at all the preset timesteps or approximation with dense predicted pose sequences~\cite{li2018convolutional}, which both involve unnecessary computations. Therefore, we need a more direct method, avoiding unnecessary overhead computations, to achieve responsive and accurate anytime pose predictions.

Recent deep learning advancements have led to continuous-time neural functions inspired by the Ordinary Differential Equations (ODEs)~\cite{rubanova2019latent,park2020vid,franceschi2020stochastic}. The functions are known as Neural ODEs and have demonstrated their effectiveness in continuous temporal sequence modeling but have not been explored in challenging tasks of unstructured, spatio-temporal human 3D pose sequence modeling and forecasting.   

Inspired by the continuous-time neural functions and the need for anytime pose forecasting in real systems, we propose AnyPose, a framework to forecast future 3D human poses at any desired time step. AnyPose is a Neural ODE-based framework that evolves a human pose to any given timestep within time bounds by solving the initial value problem (IVP). The main contributions of this paper are summarized as follows:

\begin{itemize}[leftmargin=*]
    \item Our paper highlights a need for anytime pose forecasting models and introduces the first continuous-time framework, AnyPose, to solve such tasks with high performance.
    \item We propose AnyPose, a Neural ODE-based framework for short and long-horizon anytime 3D human pose forecasting.
    \item We evaluate the accuracy and speed performance of AnyPose on various datasets and conduct a series of analytical experiments to study the intersection of human pose prediction and Neural ODE, which can serve as the first reference and analysis for future works in this field.
\end{itemize}

\begin{figure*}[ht]
  \centering
  \vspace{2.5mm}
  \includegraphics[width=1.0\textwidth]{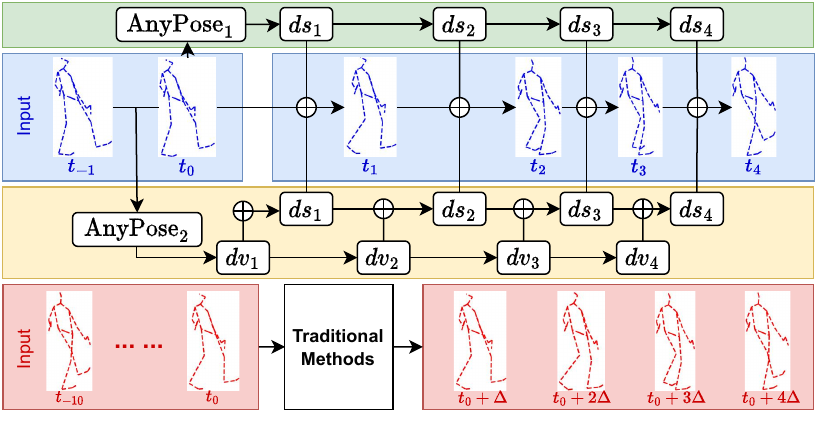}
  \caption{AnyPose takes a few (1-2) past pose sequences to predict human 3D poses at any desired time step, whereas the traditional method only forecasts poses at preset, discrete time intervals while taking a large number of past pose sequences as an input.}
  \label{fig:workflow}
  \vspace{-6mm}
\end{figure*}

\section{Related Work}
\label{sec:relatedwork}
This section presents relevant work in Human 3D pose forecasting, ODE-based continuous-time neural models, and multi-headed self-attention encoding for long-horizon sequence modeling.

\textbf{3D Human Pose Forecasting.} 
The earlier work in human pose forecasting involves traditional methods such as the hidden Markov model~\cite{lehrmann2014efficient, brand2000style}, and Gaussian process~\cite{wang2007gaussian}. Recent developments led to sequential deep learning models such as Recurrent Neural Networks (RNNs)~\cite{fragkiadaki2015recurrent, martinez2017human, walker2017pose, gopalakrishnan2019neural} and their variants with gated memory units~\cite{du2019bio, graves2012long, chung2014empirical, jaouedi2020prediction}. Although these neural frameworks outperformed traditional methods, they struggled in long-horizon human pose forecasting tasks~\cite{mao2020history}. Generative Adversarial Networks (GANs)~\cite{gui2018adversarial} were also employed with reinforcement learning to solve long-term prediction problems~\cite{wang2019imitation, qi2020imitative}. However, the quest for better frameworks evolved towards neural graph or attention-based architectures~\cite{aksan2021spatio, mao2020history, martinez2021pose, cai2020learning, sofianos2021space}, which encode spatio-temporal pose sequences for long-horizon forecasting tasks~\cite{aksan2021spatio}. These methods leverage Graph Neural Networks (GNNs) for spatial encoding and Convolutional Neural Networks (CNNs) or Temporal Convolutional Networks (TCN) for temporal encodings of human pose sequence modeling and forecasting~\cite{butepage2017deep, cao2020long, hernandez2019human, holden2015learning, bai2018empirical, holden2015learning, gehring2017convolutional}. Among these methods,~\cite{mao2020history} leverage the attention architecture to actively search patterns of periodic actions from observed human pose trajectories and demonstrates state-of-the-art performance on various standard human activity datasets, including Human3.6M, AMASS~\cite{AMASS:ICCV:2019}, and 3DPW~\cite{vonMarcard2018}. However, these methods forecast human poses at preset, discrete-time steps and do not evolve human postures in continuous time-space for any time forecasting tasks.

\textbf{Neural ODEs} Another relevant area to our work is the continuous-time neural architectures formulated by ODEs. The Neural ODE~\cite{chen2018neural} introduced ODE-based hidden neural layer evolution with applications ranging from continuous normalizing flows to generative time-series modeling. The adjoint sensitivity method allowed gradient computation through ODE solvers for backpropagation.~\cite{rubanova2019latent} proposed RNN-ODE, an ODE-based RNN model, which is capable of modeling irregularly sampled data points. The second-order Neural ODEs~\cite{yildiz2019ode2vae} were also introduced based on optimal transportation theory to model complex dynamical systems such as bouncing balls from raw visual observations. However, ODEs have some special issues. For example, due to the nature of adaptive step size ODE solvers, the situation that many consecutive layers are dynamically equivalent is very common, in~\cite{finlay2020train}, the problem is solved by applying optimal transportation theory to encourage simpler trajectory dynamics. Recent developments extend these ideas to continuous-time video forecasting~\cite{park2020vid} and continuous attention architectures~\cite{liu2020learning, zhang2021continuous}. However, the application to anytime human 3D pose forecasting has not been explored. To the best of our knowledge, AnyPose is the first framework to solve such unstructured long-horizon forecasting tasks in a continuous time-space.

\section{Anytime 3D Human Pose Forecasting}
\label{sec:methods}
In this section, we present AnyPose, a neural architecture for forecasting human pose. Compared with traditional methods, AnyPose features superior flexibility as it is able to predict future human poses at any desired time instant. Within this architecture, we propose two models, AnyPose$_1$ and AnyPose$_2$, which are respectively relied on first- and second-order Neural ODE (ODE$_1$ and ODE$_2$) and in favor of different prediction horizon. We outlined their architectures in~\ref{fig:workflow}, including the neural ODE-based continuous human pose dynamics modeling, initial states formation, and the working pipelines.

\subsection{Problem Formulation}
\label{subsec:pf}
This section formulates the problem of anytime pose forecasting and its distinction from existing discrete sequential models. Let $s \in \mathbb{R}^{3\times M}$ be a $3\times M$ dimensional human body pose comprising the 3D spatial position of $M \in \mathbb{N}$ joints. Let a sequence of human poses be denoted by a $\mathcal{S}$, i.e.,  $s\in S$. The existing state-of-the-art pose forecasting methods generate a future pose sequence $\{s_i,s_{i+\Delta t}, s_{i+2\Delta t}, \cdots, s_{i+N\Delta t} \}$ in preset discrete intervals $\Delta t$ conditioned on past behavior sequences $\{s_{i-N'\Delta t},\cdots,s_{i-2\Delta t}, s_{i-\Delta t}\}$ until $N$ steps. In addition, these intervals are often implicitly incorporated through the sampling rate of training human motion trajectories. However, Anytime pose forecasting models are a function of real-valued time $t \in \mathbb{R}$ and previous pose sequences. The input time $t$ indicates the instant at which the pose is to be inferred. However, predicting pose at any real-valued time instant requires modeling human behaviors in continuous-time space. 

In the context of Neural ODE, pose prediction can be either interpreted as a first-order or second-order IVP problem. ODE$_1$ models the derivatives of spatial positions of each joint in time domain: 
\begin{gather}
\label{eq:v}
    \dot{\boldsymbol{s}}=v=\frac{d\boldsymbol{s}}{dt},
\end{gather}
namely velocity, and evolve initial poses by integrating them to the given timestep. ODE$_2$ tackles the problem in a deep layer. It models the derivatives of velocity, i.e., the second-order derivative of joints positions
\begin{gather}
\label{eq:a}
    \ddot{\boldsymbol{s}}=a=\frac{d^2\boldsymbol{s}}{dt},
\end{gather}
as human motion generally involves accelerations. In the following sections, we present both methods and their corresponding models, AnyPose$_1$ and AnyPose$_2$.

\begin{figure*}[t]
  \centering
  \vspace{2.5mm}
  \includegraphics[width=1.0\textwidth]{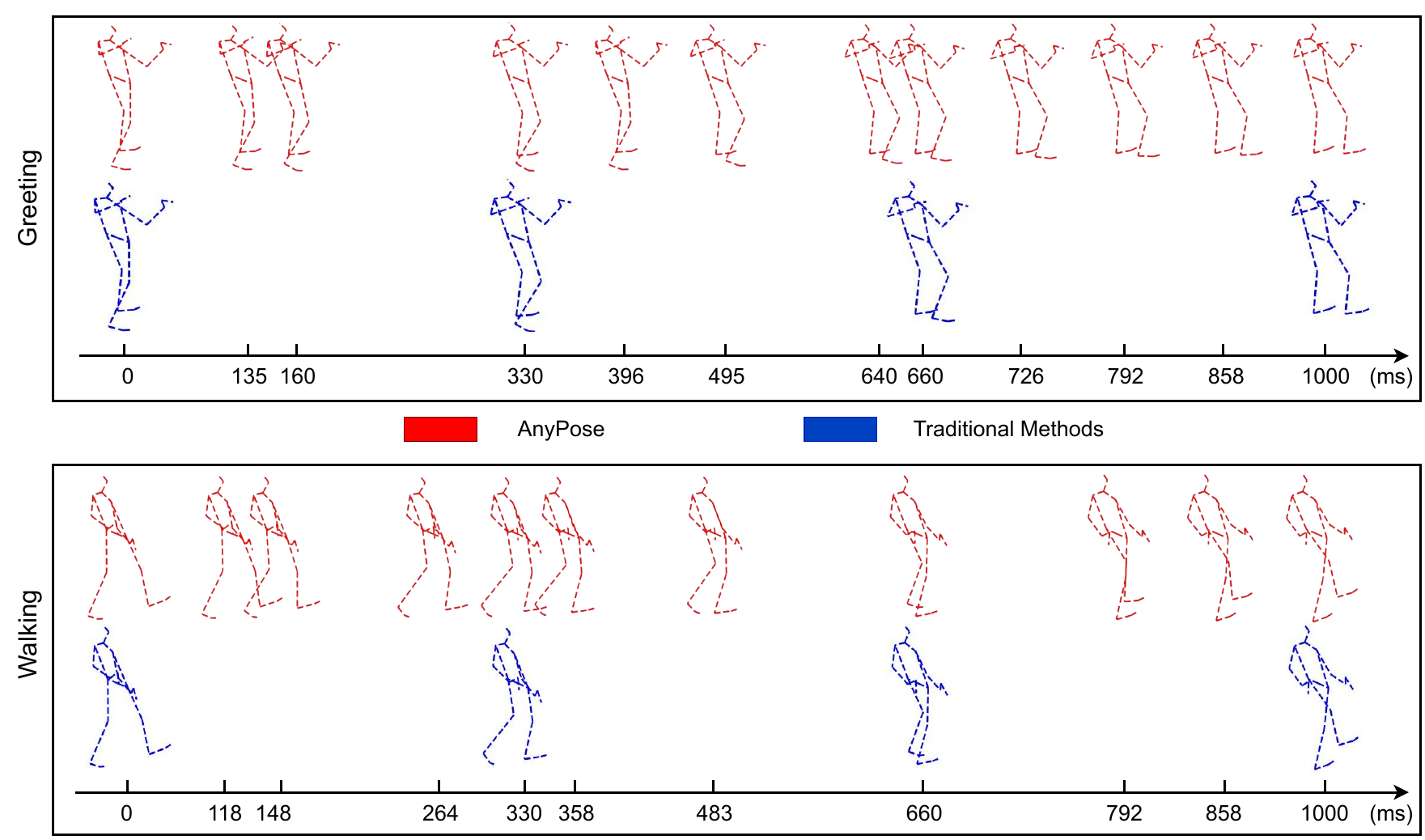}
  \caption{AnyPose models continuous time pose sequences, allowing it to forecast poses at any given time step. On the contrary, traditional methods can only predict future poses at fixed, preset time steps. (Note: Although traditional methods predict poses every 25ms, we select postures that align with the indicated time steps for illustration.)}
  \label{fig:cmp}
  \vspace{-6mm}
\end{figure*}

\subsection{First-order Neural ODE and AnyPose$_1$}
\label{subsec:ap1}
In general calculus, an initial value problem consists of an ordinary differential equation and an initial condition that specifies the value of a variable of interest $h$ at a given time $t$. In~\cite{chen2018neural}, a generative latent function time-series model, latent ODE, of neural ODEs family is proposed to solve first-order IVP:
\begin{gather}
\label{eq:ode1}
    \frac{dh(t)}{dt} = f_{\theta}(h(t)) \quad\textrm{where}\quad h(t_0) = h_0.
\end{gather}
In this equation, the solution to this IVP at $t$ is $h(t)$, in which the $t$ is the desired time. The initial condition is defined by the initial time $t_0$ and initial value $h(t_0)$. The dynamics of the $h(t)$ is described by function $f_{\theta}$, which is a neural network with the parameter $\theta$. In addition, the solution $h(t)$ can be evaluated at any given times defined within the range using a numerical ODE solver:
\begin{gather}
\label{eq:ode1solve}
    h_0,\dots, h_N = \textrm{ODESolve}(f_\theta,~h_0,~(t_0, \dots, t_N)),
\end{gather}
where the ODESolve represents any traditional ODE solver, and the $(t_0,\cdots, t_N)$ is the list of all desired time steps at which the output solution is computed.

AnyPose$_1$ is constructed based on Eq.~\ref{eq:ode1} and Eq.~\ref{eq:ode1solve}. In the context of pose prediction, the dynamics of human motion, i.e., the first-order derivative, $\frac{dS(t)}{dt}$, 
of the human pose at any given time step $t$ 
is hard to explicitly model. Therefore, we employ a Multi-layer Perceptron (MLP$_1$) Network to perform such a task. The initial state is naturally and explicitly represented by the last observed pose, i.e., $S(0) = s_{0}$, 

which is also the input data of AnyPose$_1$. Hence, Eq.~\ref{eq:v} and Eq.~\ref{eq:ode1} can be reformed as:
\begin{gather}
\label{eq:ap1}
    V = \frac{dS(t)}{dt} = \text{MLP}_1(S(t)) \quad\textrm{where}\quad S(0) = s_0.
\end{gather}
Given Eq.~\ref{eq:ap1} and integration rules, AnyPose$_1$ predicts the future pose $s_t$ at time step $t$ as follow:
\begin{gather}
\label{eq:ap1-integrate}
    s_t = s_0 + \int_{0}^{t}\text{MLP}_1(S(t)).
\end{gather}
Since integration is a sequential process, the poses at all the desired time steps can be computed within one execution of ODE Solver:
\begin{gather}
\label{eq:ap1-odesolve}
    s_1,\dots, s_N = \textrm{ODESolve}(\text{MLP}_1, s_0,(t_1, \dots, t_N)),
\end{gather}
It is worth noting that the original design of latent ODE encodes input data to latent space with neural networks and uses it as the initial state under the variational auto-encoder structure (VAE). Unlike it, we directly use raw pose data, which does not need an encoder and decoder module, since we found such a way yields a better result (Refer to~\ref{subsec:ae} for more details).

\subsection{Second-order Neural ODE and AnyPose$_2$}
\label{subsec:ap2}
As discussed in section~\ref{subsec:pf}, the pose forecasting problem can not only be formulated with human motion velocity, but also acceleration. In the belief that acceleration, which resides in the deeper layer of motion, is able to reveal more information about general dynamics, we propose AnyPose$_2$ working as follows.

Similar to the first order IVP described in~\ref{subsec:ap1}, we can extend Eq.~\ref{eq:ode1} with Eq.~\ref{eq:a} to represent the second order IVP as:
\begin{gather}
\label{eq:ode2}
    \frac{d^{2}h(t)}{dt^2} = f_{\theta}(h(t),~\Dot{h}(t)) \quad \nonumber \\ 
    \textrm{where}\quad h(t_0) = h_0,~\Dot{h}(t_0) = \Dot{h}_0.
\end{gather}
The ODE Solver get the solution $h(t)$ and $\Dot{h}(t)$ by integrating with $f_{\theta}$:
\begin{gather}
\label{eq:ode2solve}
    (h_1,~\Dot{h}_1),\dots, (h_N,~\Dot{h}_N) = \nonumber \\ 
    \textrm{ODESolve}(f_\theta,~(h_0,~\Dot{h}_0),~(t_0,\dots, t_N)).
\end{gather}


In the context of AnyPose$_2$, the initial state modeling is more complicated. As Eq.~\ref{eq:ap1} indicates,~$h(t)$ can be represented with the last observed pose $s_0$. In terms of~$\Dot{h}(t)$, by definition, it should be the velocity $v_0$ of each joint of $s_0$. Thus, we obtain $v_0$ from last two observed poses ($s_{-1}$, $s_0$) with inter-frame time interval: $v_0 = (s_0 - s_{-1}) / \Delta t$. Along with a MLP for AnyPose (MLP$_2$) and Eq.~\ref{eq:a},~\ref{eq:ode2} is transformed into:
\begin{gather}
\label{eq:ap2}
    A = \frac{d^{2}S(t)}{dt^2} = \text{MLP}_2(S(t),~\Dot{V}(t)) \quad \nonumber \\ 
    \textrm{where}\quad S(0) = s_0,~V(0) = v_0.
\end{gather}
Following~\cite{yildiz2019ode2vae}, we can reduce it into two coupled first-order ODEs to represent the velocity and acceleration models of human behavior, i.e.,
\begin{gather}
\label{eq:ode2_dc}
    \left[
    \begin{aligned}
        s_t\\v_t
    \end{aligned}
    \right]
    =
    \left[
    \begin{aligned}
        s_0\\v_0
    \end{aligned}
    \right]
    +
    \int_{0}^{T}
    \left[
    \begin{aligned}
        v_t&\\~\text{MLP}_2(S(t),&~\Dot{V}(t))
    \end{aligned}
    \right] dt
\end{gather}
As discussed in~\cite{yildiz2019ode2vae}, s$_t$ is defined on the velocity variable. Thus acceleration serves as an auxiliary variable driving the velocity forward with integration progress. Similar to Eq.~\ref{eq:ap1-odesolve}, AnyPose$_2$ effectively outputs poses at desired time steps in one run.

\subsection{Objective Function}
\label{subsec:supv}
We used the standard Mean Per Joint Position Error (MPJPE) as a loss function between the ground truth and predicted poses at desired time steps:
\begin{align}
\label{eq:loss}
    \frac {1}{M} \sum _{j=1}^{M} ||\hat {\boldsymbol {x}}_{j}-\boldsymbol {x}_{j} ||_{2}.
\end{align}
The $M$ denotes the total number of joints of a human pose, $\hat {\boldsymbol {x}}_{j} \in \mathbb{R}^3$ represents the 3D coordinates of the joint $j$ of the predicted pose and ${\boldsymbol {x}}_{j} \in \mathbb{R}^3$ represents the ground truth corresponding to $\hat {\boldsymbol {x}}_{j}$. Though AnyPose generates continuous pose trajectories, we only calculate the MPJPE on poses sampled at the same time point as ground truth.

\begin{table*}[ht]
\caption{Full-term Average MPJPE on Human3.6m Dataset}
 \begin{center}
 \vspace{-3mm}
 \resizebox{0.7\textwidth}{!}{
\begin{tabular}{c|cccccccc}

\toprule
msec                                    & 80    & 160   & 320   & 400   & 560   & 720    & 880   & 1000   \\ \midrule
Res. Sup.~\cite{martinez2017human}      & 25.0  & 46.2  & 77.0  & 88.3  & 106.3 & 119.4  & 130.0 & 136.6  \\
ConvSeq2Seq~\cite{li2018convolutional}  & 16.6  & 33.3  & 61.4  & 72.7  & 90.7  & 104.7  & 116.7 & 124.2  \\
STS-GCN~\cite{sofianos2021space}        & 17.7  & 33.9  & 56.3  & 67.5  & 85.1  & 99.4   & 109.9 & 117.0  \\ 
Motion Attention~\cite{mao2020history}  & 10.4  & 22.6  & 47.1  & 58.3  & 77.3  & 91.8   & 104.1 & 112.1  \\ \midrule
L-AnyPose$_1$                           & 104.3 & 115.5 & 121.3 & 127.8 & 132.9 & 133.6 & 139.1 & 140.3 \\
L-AnyPose$_2$                           & 52.5  & 62.9  & 82.9  & 91.4  & 105.3 & 116.1 & 125.1 & 131.7 \\ \midrule
AnyPose$_1$ (Ours)                      & 22.8  & 41.9  & 69.9  & 80.6  & 97.5  & 110.2  & 121.2 & 128.2  \\
AnyPose$_2$ (Ours)                      & 15.9  & 31.5  & 68.0  & 84.0  & 108.4 & 123.1  & 133.3 & 139.3  \\ \bottomrule
\end{tabular}}
\end{center}
 
 \label{table:avg-mpjpe}
 \vspace{-6mm}
\end{table*}

\section{Experiments}
The objectives of our experiments are twofold. Firstly, we evaluate the accuracy and speed performance of AnyPose$_1$ and AnyPose$_2$. Secondly, we conduct a series of modular experiments to comprehensively analyze the effect of different architectures in the intersection of Neural ODE and human poses. Through this section, we target to provide baseline performance and the first-ever analysis of the intersection of Neural ODE and human pose forecasting.

\subsection{Dataset and Metric}
\textbf{Human3.6M} dataset~\cite{ionescu2013human3}, commonly used in various human pose estimation and prediction tasks. It consists human poses captured from 7 subjects (actors), performing 15 ordinary actions such as walking, smoking, and sitting. Each pose in the dataset is represented by 32 joints, which are computed by forward kinematics~\cite{fragkiadaki2015recurrent} from an exponential map of joints. To reduce the redundant and unnecessary points, we only consider 22 joints out of the provided 32 in the training and testing process. We use the established train, validation, and test split following the previous work~\cite{mao2019learning, sofianos2021space}, which assigns subject 11 (S11) as the validation set, subject 5 (S5) for the testing set, and all other subjects for the training set.

\textbf{AMASS}, the Archive of Motion Capture as Surface Shapes dataset~\cite{AMASS:ICCV:2019} is a collection of a series of mocap datasets with the unified representation of SMPL~\cite{SMPL:2015}, which represents a human by a shape vector and joint rotation angles. Following~\cite{mao2020history}, we use BMLrub2 as the test set and convert the remaining parts of AMASS into training and validation data.

\textbf{3DPW}, 3D Pose in the Wild dataset~\cite{vonMarcard2018} is a large-scale dataset including indoor and outdoor actions. Following~\cite{mao2020history}, we use the test set of 3DPW to evaluate the AnyPose trained with AMASS data to check the generalization ability of our approach.

To evaluate the performance of AnyPose, we use the commonly used metric MPJPE. Though mean angle error (MAE), which computes the Euler angle error between the predicted pose and its corresponding ground truth pose, is also widely used, it suffers from an inherent ambiguity~\cite{sofianos2021space}, which makes it less effective than MPJPE. Therefore, we only adopt MPJPE as our evaluation metric. The detail of MPJPE is well-described in~\ref{subsec:supv}.

\begin{table*}[ht]
\caption{Full-term MPJPE on AMASS and 3DPW Dataset}
 \begin{center}
 \vspace{-3mm}
 \resizebox{1.0\textwidth}{!}{
 \begin{tabular}{c|cccccccc|cccccccc}
    \toprule
    Dataset &\multicolumn{8}{c}{AMASS-BMLrub} &\multicolumn{8}{c}{3DPW} \\ \midrule
    msec                                    & 80    & 160   & 320   & 400   & 560   & 720   & 880   & 1000  & 80    & 160   & 320   & 400   & 560   & 720   & 880   & 1000 \\ \midrule
    Res. Sup.~\cite{martinez2017human}      & 20.6  & 39.6  & 59.7  & 67.6  & 79.0  & 87.0  & 91.5  & 93.5  & 18.8  & 32.9  & 52.0  & 58.8  & 69.4  & 77.0  & 83.6  & 87.8 \\
    convSeq2Seq~\cite{li2018convolutional}  & 20.6  & 39.6  & 59.7  & 67.6  & 79.0  & 87.0  & 91.5  & 93.5  & 18.8  & 32.9  & 52.0  & 58.8  & 69.4  & 77.0  & 83.6  & 87.8 \\
    STS-GCN~\cite{sofianos2021space}        & 13.6  & 25.3  & 44.6  & 52.7  & 61.2  & 72.4  & 78.6  & 84.0  & 14.3  & 25.4  & 42.8  & 49.3  & 59.6  & 67.3  & 72.6  & 76.3 \\
    Motion Attention\cite{mao2020history}   & 11.3  & 20.7  & 35.7  & 42.0  & 51.7  & 58.6  & 63.4  & 67.2  & 12.6  & 23.1  & 39.0  & 45.4  & 56.0  & 63.6  & 69.7  & 73.7 \\ \midrule
    AnyPose$_1$ (Ours)                      & 18.5  & 26.5  & 55.7  & 63.8  & 69.8  & 76.6  & 85.4  & 91.7  & 21.9  & 39.2  & 53.8  & 59.7  & 69.1  & 75.9  & 80.8  & 84.4 \\ 
    AnyPose$_2$ (Ours)                      & 15.1  & 24.5  & 49.9  & 58.1  & 68.2  & 78.0  & 88.1  & 97.6  & 16.1  & 29.5  & 41.1  & 55.3  & 65.3  & 78.9  & 84.8  & 93.3 \\ 
    \bottomrule
 \end{tabular}}
 \end{center}
  \vspace{-5mm}
  \label{table:othersets}
\end{table*}

\subsection{Baselines}
As AnyPose is the first Neural ODE-based pose forecaster, we choose the representative or advanced method of the following mainstream architectures as the baselines of our experiments: Res. sup.~\cite{martinez2017human} for recurrent neural network (RNN), ConvSeq2Seq~\cite{li2018convolutional} for convolutional neural network (CNN), STS-GCN~\cite{sofianos2021space} for convolutional graph neural network (GCN) and Motion Attention~\cite{mao2020history} for attention (ATTN).

\subsection{Accuracy Evaluation}
\label{subsec:ae}
We evaluate the prediction accuracy of AnyPose on Human 3.6M, AMASS, and 3DPW datasets and compare the performance with baselines. The evaluated methods are tasked to predict the next 2, 4, 8, and 10 future poses in the short-term evaluation and 14, 18, 22, and 25 future poses in the long-term evaluation. The baseline methods take the last 10 observed poses to predict short- and long-term future poses. In contrast, AnyPose$_1$ takes only 1 previously observed pose, and AnyPose$_2$ takes only the previously observed 2 poses for short- and long-term predictions.  

The results are reported in Tab.~\ref{table:avg-mpjpe} and~\ref{table:othersets}. In summary, it can be seen that AnyPose methods achieved comparable results as state-of-the-art baseline methods. However, in contrast to baseline methods which require the last 10 poses, the AnyPose methods take only 1-2 previous poses for making future predictions. Thus, it can be concluded that AnyPose is an ideal instant pose forecasting approach as, unlike traditional methods, it does not rely on too many past observed poses for making comparable future pose predictions.

Furthermore, in a relatively short time horizon, AnyPose shows a similar level of performance as Res. Sup. and ConvSeq2Seq, whereas the GCN and ATTN-based methods demonstrate better accuracy. However, compared to Res. Sup. and ConvSeq2Seq, the neural network structure of AnyPose is much smaller and simpler. We conjecture that the reason for good AnyPose performance with only a few input past poses is that the integration flow of ODE is the more suitable method to interpret temporal or sequential relationships in pose forecasting. Moreover, the adaptive size of integration is able to help more accurately catch the evolving trajectories of human poses. In terms of STS-GCN and Motion Attention, our analysis assumes their performance advantage over other methods is mainly brought by the spatial encoding ability of the joint-wise graph~\cite{sofianos2021space} or the attention of overlapped pose group~\cite{mao2020history}, whereas other methods including AnyPose are primarily aiming to model the temporal dynamics. The result of the experiment related to our proposed trajectory optimizer in section~\ref{subsec:as} echos our assumption.

In a long time horizon, which is the most challenging part of the human prediction task, the relatively simple structure of AnyPose can still capture the dynamics of human motion. The overall performance distribution of all the methods is identical to the short time horizon. However, we noticed that one of the reasons that long-horizon prediction becomes challenging for all methods is mode switching. For instance, human motion sequences, as time evolves, might switch from walking to sitting behavior. Since pose forecasters can only rely on past observed poses to make predictions, none of the presented and baseline methods can infer the upcoming novel motion in the long horizon, which is reflected by the increased MPJPE. 

In our supplementary material, we also present a detailed action-wise MPJPE comparison of our methods with the baselines to highlight the action sequences causing challenges in solving prediction tasks.


\subsection{Speed Evaluation}
\label{subsec:se}
The lightweight model and ODE integration flow endow AnyPose with an unparalleled speed advantage. To quantitatively evaluate the speed of AnyPose and other baselines, we uniformly sampled 1000 time steps within 1 second with corresponding test pose sequences randomly drawn from the test set of Human 3.6M, and the evaluated methods are tasked to predict the pose at the given time either use ODE solver or joint-wise linear interpolation, a few examples shown in~\ref{fig:cmp}. The experiment is conducted on a desktop computer with an NVIDIA RTX 3090 GPU and Intel Core i7-11700k processor.

We report the result in Tab.~\ref{table:speed}. As the results show, AnyPose methods demonstrate an irrefutable inference speed advantage. Given our testing GPU specification, the significant time gap will be further amplified in more realistic devices such as edge computing nodes and CPU-only platforms. We believe the source of the speed advantage is clear. Firstly, comparing the large neural networks, especially attention networks, AnyPose depends on a very lightweight MLP, which makes the module contribute the most to speed as small networks require limited computations. Secondly, because of the sequential execution behavior and the flexibility brought by the ODE solving scheme, AnyPose is able to fast march to the desired timestep and terminate immediately after the pose is generated. However, baseline methods must infer poses at all the preset time step and then interpolate them to achieve the target. Such a working scheme costs a lot of unnecessary computational time, especially when the desired time is on the short-term horizon. 
Finally, AnyPose does not involve the input encoding and latent embedding decoding process as discussed in~\ref{subsec:ap1}. Instead, it directly uses the raw pose to start the inference, which also saves a great amount of time. It is worth noting Res. Sup. can be executed in auto-regressive behavior, which can partially achieve the early termination feature of AnyPose, but the interpolation and the encoding-decoding process are unavoidable. In conclusion, we believe AnyPose achieves our design target of fast inference speed, which makes it suitable for many time-sensitive scenarios such as smooth HMI.

\begin{table}
\caption{Anytime Inference Speed Evaluation}
\begin{center}
\vspace{-3mm}
\resizebox{0.37\textwidth}{!}{
    \begin{tabular}{c | c c}
    \toprule
                     & Mean Time Cost & Variance \\ \midrule
    Res. Sup.        & 5.81e-1 sec    & 3.6e-6   \\
    ConvSeq2Seq      & 3.11e-2 sec    & 1.1e-5   \\
    STS-GCN          & 8.47e-2 sec    & 7.9e-8   \\
    Motion Attention & 9.01e-1 sec    & 2.6e-7   \\ \midrule
    AnyPose$_1$      & 1.99e-3 sec    & 1.3e-4   \\
    AnyPose$_2$      & 2.64e-3 sec    & 1.9e-4   \\ \bottomrule
    \end{tabular}}
\end{center}
\label{table:speed}
\vspace{-6mm}
\end{table}

\subsection{Proposed Framework Analysis}
\label{subsec:as}
This section investigates the impact of the different modules and architectures to AnyPose design on its overall performance. The results related to our discussion are summarized in the Tab.~\ref{table:avg-mpjpe} and~\ref{table:othersets}.

\textbf{Latent Neural ODE}. According to the original design of Neural ODE~\cite{chen2018neural} and Neural ODE$_2$~\cite{yildiz2019ode2vae}, the trajectory evolvement is conducted in latent space, which requires neural networks to encode the input sequence and decode the latent output embeddings into pose space. We choose to retain our AnyPose method in pose space because our experiments show that latent ODE does not work well in pose prediction tasks.
We also observed that in some cases the latent neural ODE-based pose forecasting cannot output legit human-like poses. To validate our claim, we provide the results of L-AnyPose$_1$ and L-AnyPose$_2$, which are first- and second-order latent ODE-based variants of AnyPose. In these latent AnyPose methods, the input pose  sequences, comprising the past 10 observed poses, are encoded by a Transformer network. The neural ODE module evolves the transformer-based latent encodings, representing initial conditions, and outputs the latent representation of future poses. The decoder, which is an MLP network, takes the neural ode outputs and decodes them into pose sequences. We chose Transformer~\cite{vaswani2017attention} as an encoder as it applies cross-attention to capture temporal dynamics and we observed that it leads to relatively better performance than traditional RNN-based encoders. Tab.~\ref{table:avg-mpjpe} compares the performance of L-AnyPose variants with our proposed AnyPose frameworks. It can be seen that the L-AnyPose variants exhibit inferior performance than our methods, thus validating our design choices.

\textbf{Order of ODE}. In~\ref{subsec:ap2} we state that our choice of ODE${_2}$ for AnyPose$_2$ is inspired by the fact that human motion generally involves acceleration. Thus we conjecture acceleration may reveal more information than velocity. According to the results from Tab.~\ref{table:avg-mpjpe} and~\ref{table:othersets}, we found the conjecture is not fully supported. The result shows AnyPose$_2$ does have better performance in the short-term horizon, but the situation is reversed in the long-term horizon. A possible explanation relates to the discussion in~\ref{subsec:ae}, i.e., long-horizon tasks may include mode switching. As AnyPose$_2$ has better short-term performance, it can better follow the dynamics encoded in the input pose than AnyPose$_1$, which makes it less likely to switch to the dynamics of novel and diverged motions during inference.




\section{Conclusion}
We propose AnyPose, a continuous-time neural framework for forecasting human 3D poses to any desired, real-valued time. We validate our framework on various datasets with results supporting that AnyPose predicts anytime future poses with comparable accuracy and significantly faster inference speed. We also conducted a comprehensive analysis of the intersection of Neural ODE and pose forecasters and indicated promising developing directions. Although our results indicate the original latent ODE design is unsuitable for pose prediction tasks, we still believe there is a chance for future works to successfully bring more complicated spatial and temporal encoding and decoding to the AnyPose framework.

\bibliographystyle{IEEEtran}
\bibliography{root}

\end{document}